\documentclass[review]{elsarticle}

\usepackage{lineno,hyperref}
\usepackage{booktabs}
\usepackage{subfigure}
\usepackage{soul}
\usepackage{color,xcolor}
\usepackage{multirow}
\usepackage{hyperref}
\usepackage{colortbl}
\usepackage{threeparttable}
\usepackage{amsmath,amsfonts}

\journal{Journal of \LaTeX\ Templates}

\begin{document}

\begin{frontmatter}

\title{Ultra-low Latency Spiking Neural Networks with Spatio-Temporal Compression and Synaptic Convolutional Block}
%\tnotetext[mytitlenote]{Fully documented templates are available in the elsarticle package on \href{http://www.ctan.org/tex-archive/macros/latex/contrib/elsarticle}{CTAN}.}

%% Group authors per affiliation:
%\author{Changqing Xu, Yi Liu, Yintang Yang}
%\address{Radarweg 29, Amsterdam}
%\fntext[myfootnote]{Since 1880.}

%% or include affiliations in footnotes:
\author[mymainaddress,mysecondaryaddress]{Changqing Xu\corref{mycorrespondingauthor}}
\cortext[mycorrespondingauthor]{Corresponding author}
\ead{cqxu@xidian.edu.cn}

\author[mysecondaryaddress]{Yi Liu}
\author[mysecondaryaddress]{Yintang Yang}

\address[mymainaddress]{Guangzhou Institute of Technology, Xidian University, Xi’an 710071, China.}
\address[mysecondaryaddress]{School of Microelectronics, Xidian University, Xi’an 710071, China.}

\begin{abstract}
Spiking neural networks (SNNs), as one of the brain-inspired models, has spatio-temporal information processing capability, low power feature, and high biological plausibility.
The effective spatio-temporal feature makes it suitable for event streams classification.
However, neuromorphic datasets, such as N-MNIST, CIFAR10-DVS, DVS128-gesture, need to aggregate individual events into frames with a new higher temporal resolution for event stream classification, which causes high training and inference latency.
In this work, we proposed a spatio-temporal compression method to aggregate individual events into a few time steps of synaptic current to reduce the training and inference latency.
To keep the accuracy of SNNs under high compression ratios, we also proposed a synaptic convolutional block to balance the dramatic changes between adjacent time steps.
And multi-threshold Leaky Integrate-and-Fire (LIF) models with learnable membrane time constants are introduced to increase its information processing capability.
We evaluate the proposed method for event streams classification tasks on neuromorphic N-MNIST, CIFAR10-DVS, DVS128 gesture datasets.
The experiment results show that our proposed method outperforms the state-of-the-art accuracy on nearly all datasets, using fewer time steps.
\end{abstract}


\end{frontmatter}

%\linenumbers

\section{Introduction}
Inspired by human neurons' working patterns, spiking neural networks (SNNs) are considered as the third generation artificial neural network\cite{maass1997networks}.
With the development of SNNs, a large range of applications have been demonstrated including image classification \cite{iakymchuk2015simplified}\cite{datta2021hyper}, video processing \cite{hinton2012improving} \cite{hu2016dvs}, posture and gesture recognition\cite{zhao2014feedforward}\cite{xu2020boosting}, voice recognition\cite{zhang2019spike} \cite{jin2018hybrid}.
Compared with traditional artificial neural networks (ANNs) which consist of static and continuous-valued neuron models, spiking neural networks (SNNs) have a unique event-driven computation characteristic that can respond to the events in a nearly latency-free and power-saving way\cite{pei2019towards}\cite{roy2019towards}, and it is naturally more suitable for processing event stream class.

To take full advantage of the event-driven advantages of spiking neural networks, neuromorphic sensors, such as DVS (Dynamic Vision Sensor)\cite{DVS128}\cite{CIFAR10DVS}, Asynchronous Time-based Image Sensor(ATIS)\cite{N-MNIST}, are usually used to transform datasets into neuromorphic datasets by encoding the time, location, and polarity of the brightness change.
However, the event streams recorded by neuromorphic sensor based cameras are usually redundant in the temporal dimension, which is caused by high temporal resolution and irregular dynamic scene changes\cite{yao2021temporal}.
This characteristic makes event streams almost impossible to be processed directly by deep spiking neural networks, which are based on dense computation.

To make neuromorphic datasets suitable for deep spiking neural networks, variable pre-processing methods are proposed.
The event-to-frame integrating method for pre-processing neuromorphic datasets is widely used. In \cite{SpikingJelly}, events are split into $N$ slices with nearly the same number of events in each slice and integrate events to frames. The event-to-frame integrating method can convert the event streams into tens of frames at most. Otherwise, the accuracy of the SNNs will drop significantly.
In \cite{xu2020boosting}, a temporal compression method is proposed which can reduce the length of event streams by shrinking the duration of the input event trains.
However, this method is only applied to the trained SNNs, which limits its potential.
In \cite{xu2021direct}, the normalized pixels of the static pictures are taken directly as the input current and multi-threshold neuron models are applied, which makes the SNNs can obtain a good performance in only two time steps.
This method is suitable for static image classification, but it is difficult to directly apply it to event stream classification.

In this paper, we proposed a spatio-temporal compression method to aggregate event streams into few time steps of synaptic current to reduce the training and inference latency.
To keep the accuracy of SNNs under high compression ratios, we also proposed a synaptic convolutional block in which a synaptic layer is applied to balance the dramatic change between adjacent time steps.
To increase the information processing capability of neuron models, parametric multi-threshold Leaky Integrate-and-Fire models is introduced in our SNNs.
We evaluate our method on neuromorphic datasets, such as N-MNIST\cite{N-MNIST}, CIFAR10-DVS\cite{CIFAR10DVS}, DVS128-gesture\cite{DVS128} and experimental results show that our method outperforms the state-of-the-art accuracy on nearly all tested datasets using fewer time-steps.

\section{Approach}
In this section, we first introduce the proposed spatio-temporal compression method and synaptic convolutional block in Sec. \ref{sec_compression} and \ref{sec_Synapticlayer}.
Then multi-threshold Leaky Integrate-and-Fire models with learnable membrane constants are introduced in Sec. \ref{sec_MLIF}.
At last, we describe the proposed network structure of SNNs. 

\subsection{Spatio-temporal compression method\label{sec_compression}}
\begin{figure}[t]
\centerline{\includegraphics[width=\columnwidth]{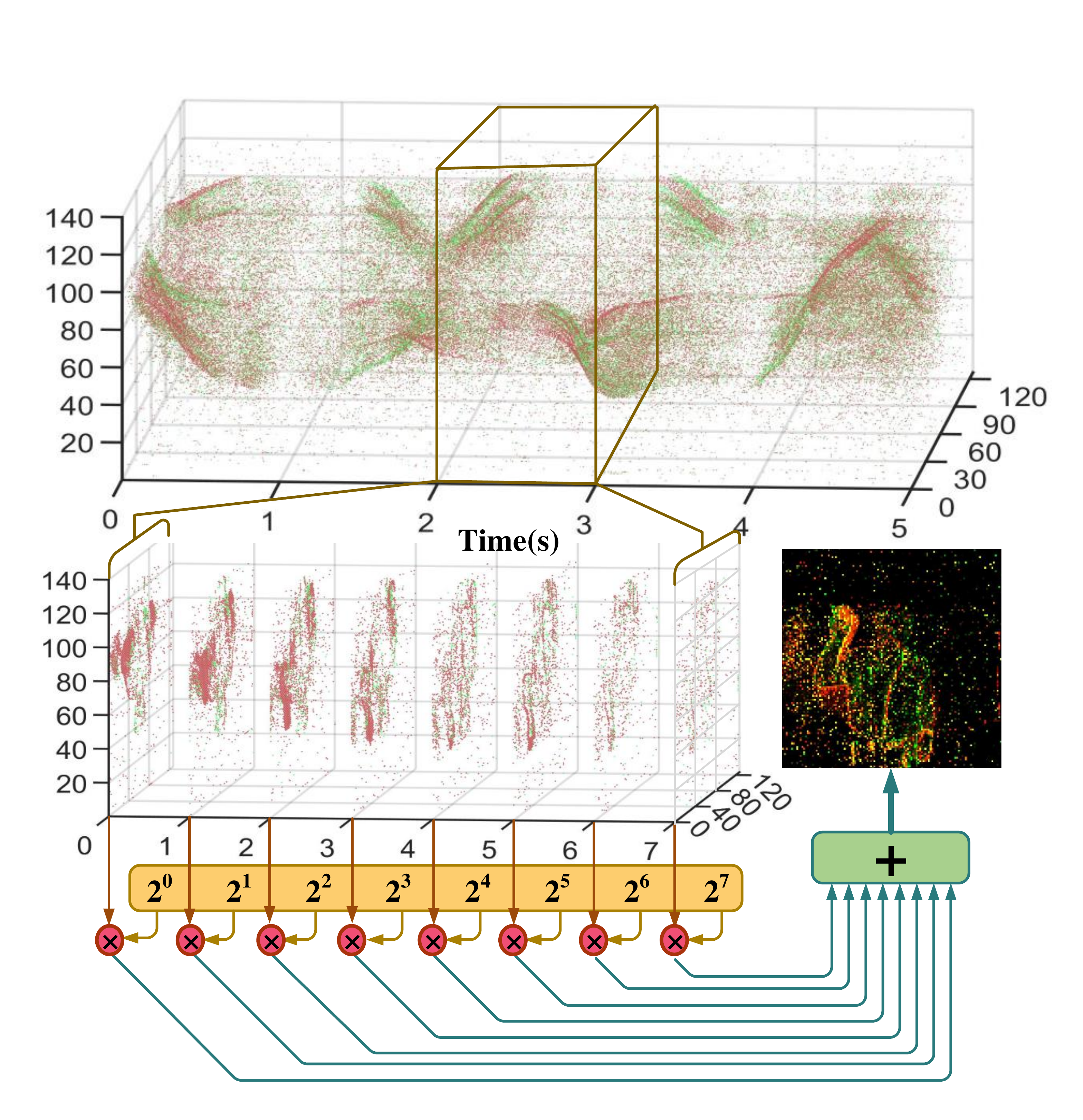}}
\caption{spatio-temporal compression block }
\label{fig_ST_Compression}
\end{figure}

The spatio-temporal compression block, which is shown in Fig. \ref{fig_ST_Compression}, is used to pre-processing neuromorphic datasets, such as N-MNIST\cite{N-MNIST}, DVS128 Gesture\cite{DVS128}, CIFAR10-DVS\cite{CIFAR10DVS}, etc.
neuromorphic datas are usually in the formulation of $E(x_{i}, y_{i}, t_{i}, p_{i})$ that represent the event's coordinate, time and polarity.
We split the event's time $T_{event}$ into $T$ slices with nearly the same time interval in each slice and integrate these events.
Note that $T$ is also the simulating time-step.
As Fig.\ref{fig_ST_Compression} shows, in each slice, spiking events are evenly divided into $N_{r}$ parts in the temporal dimension.
$N_{r}$ is the resolution that user can change based on their requirement. 
%In this paper, $N_{r}$ is set to 8.
Those spiking events in the same part will be integrated firstly and multiplied by the weights as the formulation \eqref{eq1} is shown.
Due to the spiking events having two polarities, two channels frame is used to represent the compressed neuromorphic data.

\begin{equation}
\begin{split}
&j_{l}=floor(\frac{max(T_{event})}{T})\cdot j\\
&j_{u}=\left \{
\begin{array}{cl}
    floor(\frac{max(T_{event})}{T})\cdot (j+1), & j<T-1 \\
    max(T_{event}), & j=T-1
\end{array}
\right.\\
&F(j,p,x,y)=\sum\limits_{k=0}^{N_r-1}(2^k(\sum\limits_{i=j_{l}+floor(k\frac{j_u-j_l}{N_r})}^{j_{l}+floor((k+1)\frac{j_u-j_l}{N_r})}I_{E(x,y,t,p)}(x_{i}, y_{i}, t_{i}, p_{i})))
\end{split}
\label{eq1}
\end{equation}

where $j_l$ and $j_u$ are the lower and upper bounds of the $j_{th}$ slices, respectively,
$floor()$ is the function that rounds the elements to the nearest integers towards minus infinity and  $I_{E(x,y,t,p)}(x_{i}, y_{i}, t_{i}, p_{i})$ is an indicator function of the event which equals to 1 only when $(x,y,t,p) = (x_i,y_i,t_i,p_i)$

\subsection{Synaptic convolutional block\label{sec_Synapticlayer}}

\begin{figure}[t]
\centerline{\includegraphics[width=16cm]{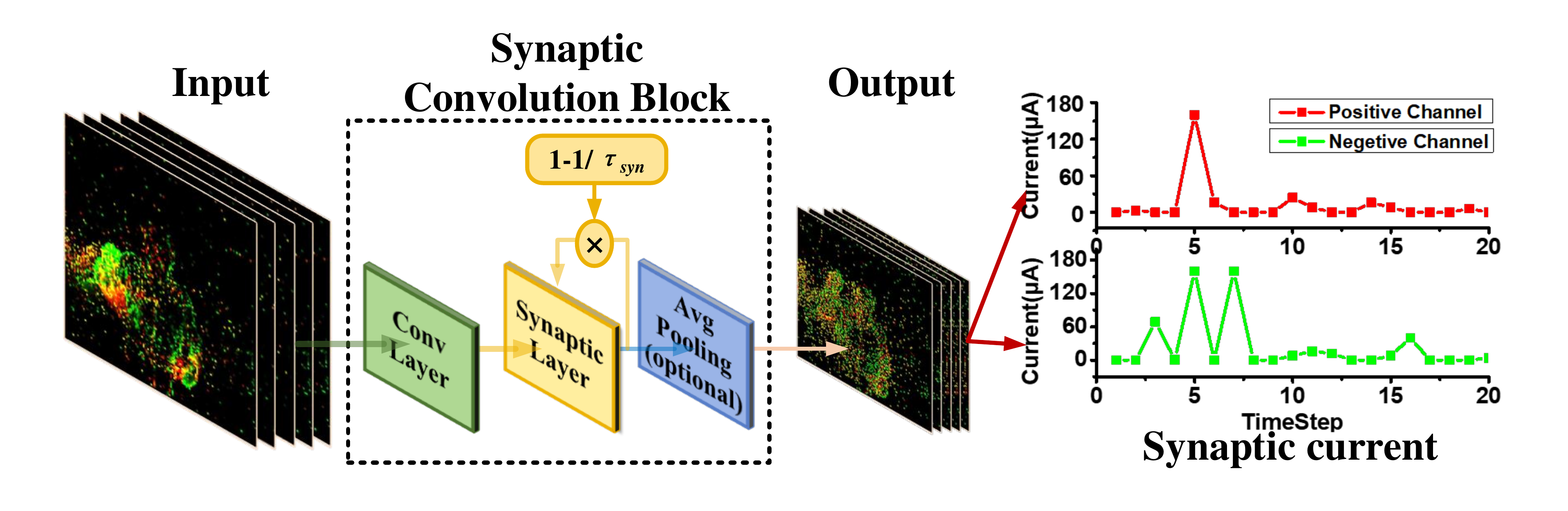}}
\caption{Synaptic convolution block}
\label{SynapticLayer}
\end{figure}

Since the synaptic convolution layer is used to balance the dramatic change between the adjacent time steps of the compressed neuromorphic data, the synaptic convolution block is used to replace the first convolution layer.
The synaptic convolution block consists of a convolution layer, a synaptic layer and a optional average pooling layer, which is shown in Fig. \ref{SynapticLayer}.
Whether there is a average pooling layer dependents on the network structure.

The key of the synaptic convolutional block is the synaptic layer.
In the synaptic layer, the first-order synaptic model is applied, which is shown below.

\begin{equation}
    I_{syn}(t)=e^{- \frac{1}{\tau_{syn}}} I_{syn}(t-1) + I_{in} (t)
    \label{eq2}
\end{equation}
where $I_{syn}(t)$ is the output current of the synapse at time $t$, $\tau_{syn}$ is the time constant of synapse and $I_{in}$ is the input current at time $t$. To facilitate the calculation and simulation, we convert Eq. \eqref{eq2} to
\begin{equation}
    I_{syn}[t_k]=(1-\frac{1}{\tau _{syn}}) I_{syn}[t_{k-1}] + I_{in} [t_k]
    \label{eq3}
\end{equation}
where $I_{syn}[t_k]$ represents the output current of the synapse at time step $t_k-1$.
$\tau _{syn}$ can control how much synaptic current at time step $t_{k-1}$ can be retained at the time step $t_{k}$.
To make the synaptic layer adjust adaptively according to the characteristics of datasets, $\tau_{syn}$ can be calculated based on 
\begin{equation}
\tau _{syn}[t_k]=\frac{1}{1-\frac{C_{valid}[t_k]}{C_{total}}}
    \label{eq3_1}
\end{equation}
where $C_{valid}[t_k]$ denotes the number of channels whose synaptic current is not zero at time step $t_k$.
$C_{total}$ is the total number of channels.
As Eq. \eqref{eq3_1} is shown, the higher the value of $C_{valid}[t_k]$ is, the higher the value of $(1-\frac{1}{\tau_{syn}})$ is, which means more information at time step $t_{k-1}$ will be retained at time $t_k$.

\subsection{Parametric Multi-threshold Leaky Integrate-and-Fire models(PMLIF)\label{sec_MLIF}}

It is known that Leaky Integrate-and Fire (LIF) model is one of the  most widely applied models to describe the neuronal dynamics in SNNs.
In this paper, we introduce the parametric multi-threshold Leaky Integrate-and-Fire models (PMLIF), whose membrane constants are learnable, to increase the information processing capability of neuron models.
The neuronal membrane potential of neuron $i$ at time $t$ is 

\begin{equation}
{\tau _m}\frac{{d{u_i}(t)}}{{dt}} =  - {u_i}(t) + I(t) + {u_{reset}}(t)
\label{eq4}
\end{equation}

Where $I(t)$ is the pre-synaptic input current at time $t$ and $\tau_m$ is a time constant of membrane voltage. $u_{reset}(t)$ denotes the reset function, which reduces the membrane potential by a certain amount $V_{th}$ after the neuron $i$ fires.
The pre-synaptic input $I(t)$ is given by

\begin{equation}
I(t) = \sum\limits_{j = 1}^N {{\omega _{ij}}s_j(t)} 
\label{eq5}
\end{equation}

Where $s_j(t)$ is the output spike of pre-synaptic neuron $j$ at time $t$ and $\omega_{ij}$ is the presynaptic weight from the neuron $j$ in the pre-synaptic layer to the neuron $i$ in the post-synaptic layer.
Due to the discrete time steps in the simulation, we apply the fixed-step first-order Euler method to discretize \eqref{eq6} to

\begin{equation}
{u_i}[t] = (1 - \frac{1}{{{\tau _m}}}){u_i}[t - 1] + I[t] + {u_{reset}}[t]
\label{eq6}
\end{equation}

Where $u_{reset}[t]$ is equal to $-s_i[t]V_{th}$ and $s_i[t]$ is the output spike of neuron $i$. In this paper, we extend the LIF model into multi-threshold LIF model, in which the output of the neuron $i$ can be expressed by

\begin{equation}
%\resizebox{.9\hsize}{!}{
%$
{s_i}[t] = \left\{ {\begin{array}{*{20}{r}}
{0,}&{{u_i}[t] < {V_{th}}}\\
{floor(\frac{{{u_i}[t]}}{{{V_{th}}}}),}&{{V_{th}} \le {u_i}[t] < {S_{max}}{V_{th}}}\\
{{s_{\max }},}&{{u_i}[t] \ge {S_{max}}{V_{th}}}
\end{array}} \right.
%$
%}
\label{eq7}
\end{equation}

Where $S_{max}$ is the upper limit of the output spikes and $floor()$ is the function that rounds the elements to the nearest integers towards minus infinity.
Since $\tau _m$ is a learnable parameter and its value should be positive, we use a sigmoid function about synaptic time constant weight $\omega _{m_{i}}$ of the neuron $i$ to replace the $(1-\frac{1}{\tau _m})$ and the \eqref{eq6} becomes
\begin{equation}
{u_i}[t] = S(\omega _{m_{i}}){u_i}[t - 1] + I[t] + {u_{reset}}[t]
\label{eq8}
\end{equation}
where $S(\omega _{m_{i}})$ is the sigmoid function about $\omega _{m_{i}}$.

\subsection{Error Backpropagation of PMLIF}
To present the error backpropagation of PMLIF, we define the loss function $L[t_k]$ in which the mean square error for each output neuron at time step $t_k$.

\begin{equation}
L[{t_k}] = \frac{1}{2}{\sum\limits_{i = 0}^{{N_o}} {({y_i}[{t_k}] - {s_i}[{t_k}])} ^2}
\label{eq9}
\end{equation}

Where $N_o$ is the number of neurons in the output layer, $y_i[t_k]$ and $s_i[t_k]$ denotes the desired and the actual firing event of neurons $i$ in the output layer at time step $t_k$.
By combining \eqref{eq4}-\eqref{eq8}, it can be seen that loss function $L[t_k]$ is a function of presynaptic weight $\omega_{i,j}$ and synaptic time constant weight $\omega_m$.
In this paper, we use $W^{(l)}= [\omega_1^{(l)}; ... ; \omega_{N_l}^{(l)}]$ to represent the presynaptic weight matrix of layer $l$ in which $\omega_{N_l}^{(l)} = [\omega_{N_l,1}^{(l)}, \omega_{N_l,2}^{(l)}, ... ,\omega_{N_l,N_{l-1}}^{(l)}]$.
$W^{(l)}_m$ denotes the synpatic time constant weight matrix, which is equal to $[\omega _{m_{1}}^{(l)}, ... , \omega _{m_{N_{l}}}^{(l)}]$. 
The aim of error backpropagation is to update the presynaptic weight $W^{(l)}$ and synaptic time constant weight $W_m^{(l)}$ using the error gradient $\frac{\partial L[t_k]}{\partial W^{(l)}}$ and $\frac{\partial L[t_k]}{\partial W_m^{(l)}}$.
Using the chain rule, the error gradient with the respect to the presynaptic weight $W^{(l)}$ in the layer $l$ is

\begin{equation}
\frac{{\partial L[{t_k}]}}{{\partial {W^{(l)}}}} = \frac{{\partial L[{t_k}]}}{{\partial {u^{(l)}}[{t_k}]}}\frac{{\partial {u^{(l)}}[{t_k}]}}{{\partial {W^{(l)}}}} = {\delta ^{(l)}}[{t_k}]\frac{{\partial {u^{(l)}}[{t_k}]}}{{\partial {W^{(l)}}}}
\label{eq10}
\end{equation}

Where $\delta ^{(l)}[{t_k}]$ is the back propagated error of layer $l$ at time $t_k$, which is equal to $\frac{\partial L[t_k]}{\partial u^{(l)}[t_k]}$. 

\begin{equation}
%\begin{array}{l}
\begin{split}
{\delta ^{(l)}}[{t_k}] &= \frac{{\partial {u^{(l + 1)}}[{t_k}]}}{{\partial {u^{(l)}}[{t_k}]}}\frac{{\partial L[{t_k}]}}{{\partial {u^{(l + 1)}}[{t_k}]}}\\
{\rm{             }}&=\frac{{\partial {u^{(l + 1)}}[{t_k}]}}{{\partial {s^{(l)}}[{t_k}]}}\frac{{\partial {s^{(l)}}[{t_k}]}}{{\partial {u^{(l)}}[{t_k}]}}{\delta ^{(l + 1)}}[{t_k}]\\
{\rm{          }} &={({W^{(l + 1)}})^T} \frac{{\partial {s^{(l)}}[{t_k}]}}{{\partial {u^{(l)}}[{t_k}]}}{\delta ^{(l + 1)}}[{t_k}]
\end{split}
%\end{array}
\label{eq11}
\end{equation}

The key to calculate $\delta ^{(l)}[t_k]$ is to obtain $ \frac{{\partial {s^{(l)}}[{t_k}]}}{{\partial {u^{(l)}}[{t_k}]}}$. Theoretically, $s^{(l)}[t]$ is a non-differentiable function and we cannot obtain the value of $ \frac{{\partial {s^{(l)}}[{t_k}]}}{{\partial {u^{(l)}}[{t_k}]}}$ directly.
In this paper, we use an approximate curve in \cite{xu2021direct} to surrogate the original derivative function. The function of the approximate curve $f_1$ is shown below.

\begin{equation}
{f_1}(u) = \sum\limits_{i = 1}^{{S_{\max }}} {{\alpha _H}{e^{ - {{(u - i{V_{th}})}^2}/{\alpha _W}}}}
\label{eq12}
\end{equation}

where $\alpha_H$ and $\alpha_W$ determine the curve shape and steep degree.
$S_{max}$ is the upper limit of the output spikes.
From \eqref{eq10},  the second term $\frac{\partial u^{(l)}[t_k]}{\partial W^{(l)}}$ is given by

\begin{equation}
\begin{split}
\frac{{\partial {u^{(l)}}[{t_k}]}}{{\partial {W^{(l)}}}} &= S(W_m^{(l)})\frac{{\partial {u^{(l)}}[{t_{k - 1}}]}}{{\partial {W^{(l)}}}} + {s^{l - 1}}[{t_k}] - \frac{{\partial {s^{(l)}}[{t_k}]}}{{\partial W^{(l)}}}{V_{th}}\\
{\rm{  }}&=S(W_m^{(l)})\frac{{\partial {u^{(l)}}[{t_{k - 1}}]}}{{\partial {W^{(l)}}}} + {s^{l - 1}}[{t_k}] - \frac{{\partial {s^{(l)}}[{t_k}]}}{{\partial {u^{(l)}}[{t_k}]}}\frac{{\partial {u^{(l)}}[{t_k}]}}{{\partial W^{(l)}}}{V_{th}}
\end{split}
\label{eq13}
\end{equation}

From \eqref{eq13}, we can get

\begin{equation}
\frac{{\partial {u^{(l)}}[{t_k}]}}{{\partial W^{(l)}}}{\rm{ = }}\frac{{S(W_m^{(l)})\frac{{\partial {u^{(l)}}[{t_{k - 1}}]}}{{\partial {W^{(l)}}}} + {s^{l - 1}}[{t_k}]}}{{1 + \frac{{\partial {s^{(l)}}[{t_k}]}}{{\partial {u^{(l)}}[{t_k}]}}{V_{th}}}}
\label{eq14}
\end{equation}

The error gradient with the respect to the synaptic time constant weight $W_{m}^{(l)}$ can be calculated by

\begin{equation}
\frac{{\partial L[{t_k}]}}{{\partial {W_{m}^{(l)}}}} = {\delta ^{(l)}}[{t_k}]\frac{{\partial {u^{(l)}}[{t_k}]}}{{\partial {W_{m}^{(l)}}}}
\label{eq15}
\end{equation}

From \eqref{eq15}, the second term $\frac{{\partial {u^{(l)}}[{t_k}]}}{{\partial {W_{m}^{(l)}}}}$ is given by

\begin{equation}
\begin{aligned}
\frac{{\partial {u^{(l)}}[{t_k}]}}{{\partial {W_{m}^{(l)}}}}
&=\frac{\partial(S(W_m^{(l)})u^{(l)}[t_{k-1}])}{\partial W^{(l)}_m}-\frac{{\partial {s^{(l)}}[{t_k}]}}{{\partial {u^{(l)}}[{t_k}]}}\frac{{\partial {u^{(l)}}[{t_k}]}}{{\partial W_m^{(l)}}}{V_{th}}
\end{aligned}
\label{eq16}
\end{equation}

From \eqref{eq16}, we can obtain

\begin{equation}
\frac{{\partial {u^{(l)}}[{t_k}]}}{{\partial {W_{m}^{(l)}}}}
=\frac{\frac{\partial(S(W_m^{(l)})u^{(l)}[t_{k-1}])}{\partial W^{(l)}_m}}{1+\frac{\partial s^{(l)}[t_k]}{\partial u^{(l)}[t_k]}V_{th}}
\label{eq16}
\end{equation}

\subsection{Network architecture\label{structure}}

\begin{figure}[t]
\centerline{\includegraphics[width=12cm]{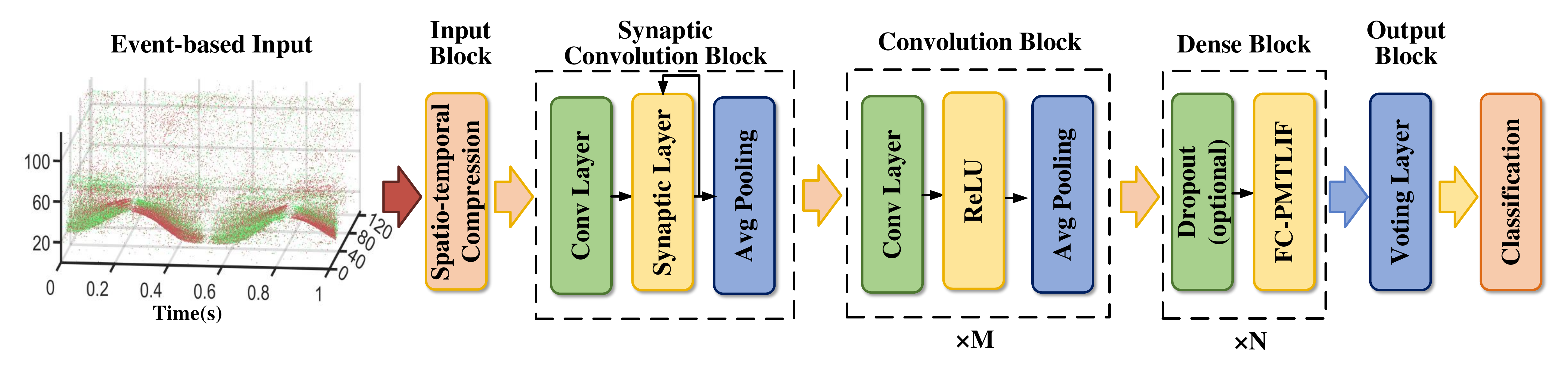}}
\caption{Architecture of the proposed SNN}
\label{Architecture}
\end{figure}

We proposed a flexible network structure for event stream classification tasks. The proposed network structure is illustrated in Fig. \ref{Architecture}.
The network consists of a spatio-temporal compression layer, a synaptic convolution block, $M$ end-to-end connected convolution blocks, $N$ end-to-end connected dense blocks, and a voting layer.
Since the output of the synaptic convolution block are synaptic currents which are real values, not binary values, we use the convolution block which consists of sequentially connected layers including a convolution layer, a ReLU layer, and an average pooling layer, to extracted features, directly.
In dense blocks, there are a dropout layer and a  fully connected layer which consists of PMLIFs.
 A voting layer after the last dense block is used to boost classifying robustness.
The output of the last dense block will be divided into $N_{class}$ groups, randomly, and connected to the voting layer, where $N_{class}$ is the number of the classes.
The voting layer is implemented by calculating the average value of each group and selecting the maximum value as the classification result.

\section{Experiments and results}
We test our proposed SNN model and training method on three neuromorphic datasets N-MNIST\cite{N-MNIST}, DVS128 Gesture\cite{DVS128}, CIFAR10-DVS\cite{CIFAR10DVS} with different sizes and structures of SNNs.
And we compared our training method with several previously reported state-of-the-art results with the same or similar networks including different SNNs trained by BP-based methods, converted SNNs, and traditional ANNs.

\subsection{Experiment settings}
\begin{table}[t]
\caption{Parameters setting}
\label{table1}
\setlength{\tabcolsep}{3pt}
\centering
\arrayrulecolor{black}
\begin{tabular}{lll} 
\hline
Parameters         & Description                               & Value                 \\ 
\hline
%\textit{$\tau_m$}        & Time constant of membrane voltage         & 10 ms                 \\
\textit{$V_{th}$}       & Threshold                                 & 10 mV                 \\
\textit{$\alpha _m$}        & Derivative approximation parameters       & 1                     \\
\textit{$\alpha _W$}        & Derivative approximation parameters       & 20                    \\
\textit{$N_{r}$}    &\begin{tabular}[c]{@{}l@{}}Resolution of spatio-temporal\\ compression\end{tabular}  & 8
\\
\textit{$S_{max}$}      & Upper limit of output spikes              & 15                    \\
\textit{$N_{Batch}$}    & \begin{tabular}[c]{@{}l@{}}Batch Size\\(N-MNIST/DVS128/CIFAR10-DVS)\end{tabular}                                & 64,16,16                   \\
\textit{$\eta$}         & \begin{tabular}[c]{@{}l@{}}Learning rate\\(N-MNIST/DVS128/CIFAR10-DVS)\end{tabular} & 0.0002, 0.00004, 0.0001  \\
\textit{$\beta _1, \beta _2, \lambda$} & Adam parameters                           & 0.9, 0.999, $1-10^{-8}$    \\
\hline
\end{tabular}
\arrayrulecolor{black}
\end{table}

All reported experiments below are conducted on an NVIDIA Tesla V100 GPU.
The implementation of our proposed method is on the Pytorch framework\cite{Pytorch}.
The experimented SNNs are based on the network structure described in Sec. \ref{structure}.
%The simulation step size is set to $1 ms$.
%Function $f_2$ is applied to approximate the derivative of spike activity.
Only 2-5 time steps are used to demonstrate the proposed ultra low-latency spiking neural network.
No refractory period is used.
Adam \cite{Adam} is applied as the optimizer.
If not otherwise specified, the accuracy in this paper refers to the best results obtained by repeating the experiments five times.

The initialization of parameters, such as the weights, threshold, time constant of membrane voltage and synapse, and other parameters, directly affect the convergence speed and stability of the whole network.
We should simultaneously make sure enough spikes transmit information between neural network layers and avoid too many spikes that reduce the neuronal selectivity.
In this paper, we use a fixed threshold in each neuron for simplification and initialize the weight $W^{(l)}$ parameters sampling from the normal distribution. 

\begin{equation}
W^{(l)}{\rm{\sim[}}\frac{{{V_{th}}}}{{{N_{l-1}}}}{\rm{,0}}{\rm{.5]}}
\label{eq17}
\end{equation}

where $V_{th}$ is the threshold of membrane voltage, $N_{l-1}$ is the number of neurons of pre-layer.
The synaptic time constant weights $W_m$ are initialized to zeros.
The set of other parameters is presented in Table \ref{table1}.
In addition, we do not apply complex skill, such as error normalization\cite{c8}, weight regularization\cite{c9}, warm-up mechanism \cite{c10}, etc.
All testing accuracy is reported after training 50 epochs in our experiments.

\begin{table}[t]
\caption{Network structure}
\label{table1_1}
    \centering
    \begin{tabular}{lc}
    \hline
    Dataset & Network structure \\
    \hline
    N-MNIST & \begin{tabular}[c]{@{}c@{}}128SC3-128C3-AP2-256C3-AP2\\-512C3-AP4-DP-512FC-10Voting\end{tabular}\\
    DVS128 & \begin{tabular}[c]{@{}c@{}}32SC3-32C3-AP2-64C3-AP2-128C3-AP2\\-256C3-AP2-512C3-AP4-DP-512FC-11Voting\end{tabular}\\
    Cifar10-DVS& \begin{tabular}[c]{@{}c@{}}32C3-32C3-AP2-64C3-AP2-128C3-AP2\\-256C3-Ap2-512C3-AP4-DP-512FC-10Voting\end{tabular} \\
    \hline
    \end{tabular}
\begin{tablenotes}
\item[1] 128SC3 represents synaptic convolution block with 128 3 $\times$ 3 filters.
128C3 represents convolution block with 128 3 $\times$ 3 filters.
AP2 represents average pooling layer with 2 $\times$ 2 filters.
DP denotes dropout layer and 512FC means a fully connected layer that consists of 512 PMLIFs.
\end{tablenotes}
\end{table}

\subsection{Dataset experiments}
\subsubsection{N-MNIST}
The Neuromorphic-MNIST (N-MNIST) dataset is a spiking version of the original frame-based MNIST dataset, which consists of the same 60 000 training and 10 000 testing samples as the original MNIST dataset\cite{N-MNIST}.
Each N-MNIST example is captured at the same visual scale as the original MNIST dataset (28x28 pixels).
The N-MNIST dataset was captured by mounting the Asynchronous Time Based Image (ATIS) sensor on a motorized pan-tilt unit and having the sensor move while it views.
For N-MNIST dataset, the network structure we applied is shown in Table \ref{table1_1}.
We compare our proposed network with several spiking convolutional neural networks which have similar network structures.
Table \ref{T1} shows that our proposed spiking neural network can achieve $99.63\%$ which outperforms other results.
In addition to the performance improvement, our proposed method also has a large reduction of time step count. 
Compared with LMCSNN\cite{LMCSNN} which only has 10 time steps, our method still has 5 times improvement.

\begin{table}[t]
\caption{Comparisons with SNNs on N-MNIST}
\label{T1}
\centering
\begin{tabular}{lllll} 
\toprule
Models      & Method & Time Step & Epoch & ACC(\%)  \\ 
\hline
TSSL-BP\cite{TSSL-BP}     & SNN    & 30         & 100   & 99.23    \\
LISNN\cite{LISNN}       & SNN    & 20        & 100   & 99.45    \\
NeuNorm SNN\cite{NeuNormSNN} & SNN    & 50        & 200   & 99.53    \\
BackEISNN\cite{BackEISNN}   & SNN    & 100       & 200   & 99.57    \\
LMCSNN\cite{LMCSNN}      & SNN    & 10        & 200   & 99.61    \\
This work   & SNN    & 2         & 50    & 99.63    \\
\bottomrule
\end{tabular}
\end{table}

\subsubsection{DVS128-gesture}
The IBM DVS128-gesture\cite{DVS128} is an event-based gesture recognition dataset, which has the temporal resolution in $\mu s$ level and 128 $\times$ 128 spatial resolution. 
It records 1342 samples of 11 gestures, such as hand clips, arm roll, etc., collected from 29 individuals under three illumination conditions, and each gesture has an average duration of 6 seconds.
Compared with N-MNIST, DVS128-gesture is a more challenging dataset which has more temporal information.
Table \ref{T2} shows that our method obtains $98.96\%$ test accuracy with only 5 time steps in 50 epochs. BPSTA-SNN \cite{BPSTA-SNN} uses the SNN which is pre-trained by TSSL-BP \cite{TSSL-BP} to obtain the same accuracy with 16 time steps in 300 epochs.

\begin{table}
\caption{Comparisons with DNNs and SNNs on DVS128 Gesture}
\label{T2}
\centering
%\resizebox{\columnwidth}{15mm}{
\begin{tabular}{llllll} 
\hline
Models     & Method & 
Time Step & Epoch & Trick                                                                             & ACC(\%)  \\ 
\hline
STBP-TDBN\cite{stbp-TDBN}  & SNN    & 40        &  -     &                                                                                   & 96.87    \\
RG-CNN\cite{RG-CNN}     & DNN    & 8         & 150   &                                                                                   & 97.20     \\
LMCSNN\cite{LMCSNN}     & SNN    & 20        & 200   &                                                                                   & 97.57    \\
STFilter\cite{STFilter}   & DNN    & 12        &  -     & \begin{tabular}[c]{@{}l@{}}Spatiotem- \\poral filters\end{tabular}                  & 97.75    \\
BPSTA-SNN\cite{BPSTA-SNN}  & SNN    & 16        & 300   & \begin{tabular}[c]{@{}l@{}}Pre-train by\\ TSSL-BP \end{tabular} & 98.96    \\
Our Method & SNN    & 5         & 50    &                                                                                   & 98.96    \\
\hline
\end{tabular}
%}
\end{table}

\subsubsection{Cifar10-DVS}
To validate our method, we apply a deeper network structure which contains six convolution layers and five average pooling, and a dense layer on the dataset Cifar10-DVS\cite{CIFAR10DVS} which is an important benchmark for comparison in SNN domains.
Cifar10-DVS is a neuromorphic version converted from the Cifar10 dataset in which 10,000 frame-based images are converted into 10,000 event streams with DVS.
Our proposed method obtains $73.8\%$ test accuracy with 5 time steps.
Compared with TA-SNN \cite{TA-SNN}, our proposed method obtain $1.8\%$ accuracy improvement with only a half of time steps.
When we apply the 10 time steps, we can achieve $76.9\%$ test accuracy in 50 epochs.

\begin{table}
\caption{Comparisons with SNNs on Cifar10-DVS }
\label{T3}
\centering
\begin{tabular}{lllll} 
\toprule
Models       & Method     & Time Step & Epoch & ACC(\%)  \\ 
\hline
NeuNorm SNN\cite{NeuNormSNN}  & SNN        & 100       & 200   & 60.5     \\
SR-ANN\cite{SR-ANN}       & ANN to SNN & 60        & -     & 66.75    \\
STBP-TDBN\cite{stbp-TDBN} & SNN        & 40        & -     & 67.8     \\
LIAF-SNN\cite{LIAF-SNN}     & SNN        & 10        & 60    & 70.4     \\
TA-SNN\cite{TA-SNN}       & SNN        & 10        & 150   & 72       \\
Our proposed & SNN        & 5         & 50    & 73.8     \\
\bottomrule
\end{tabular}
\end{table}

\subsection{Ablation Study\label{AblationStudy}}

\begin{figure}[t]
\includegraphics[width=\columnwidth]{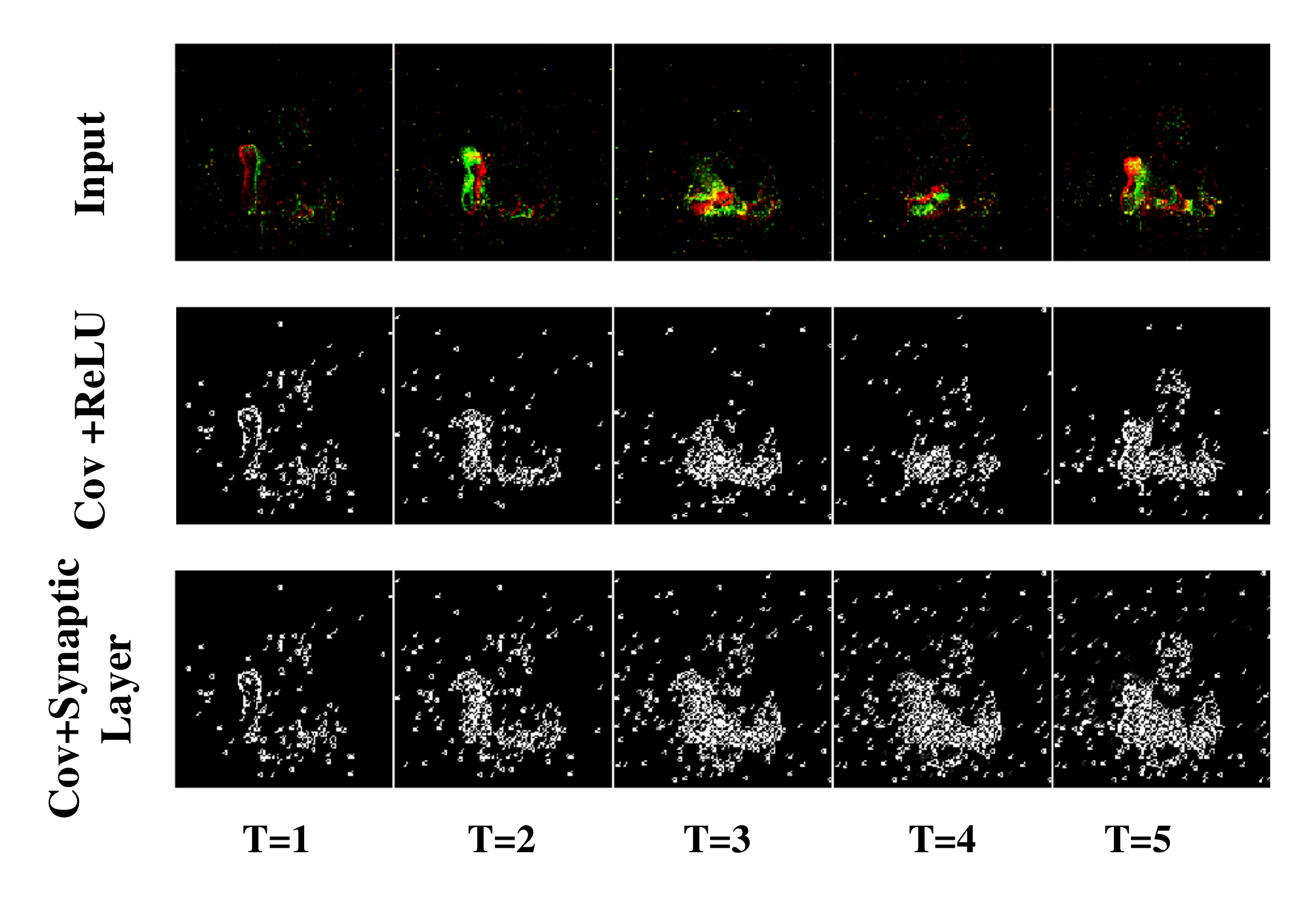}
\caption{Different between Cov+ReLU layer and Cov+Synaptic Layer }
\label{DVS128_ST}
\end{figure}

We conduct extensive ablation studies on DVS128-gesture to evaluate the validity of the method.
The reason that we chose DVS128-gesture is that it is a more challenging dataset, especially for SNNs with few time steps, due to the more complex spatial structure and stronger temporal information. 
The function of the synaptic convolutional block is to balance the dramatic change between adjacent time steps.
The dramatic change may slow down the convergence speed of the spiking neural network or even cause no converge during the training process.
We compare the output of a synaptic convolutional block with the output of a convolution layer + ReLU layer. 
In Fig. \ref{DVS128_ST}, the figures in the first row are the input data at different time steps.
Compared with figures in the second row, the figures in the third row are more continuous.
It is shown that the main difference between Cov+ReLU and Cov+Synaptic layer is that the synaptic layer retains some information of the previous time step which avoids the dramatic changes between adjacent time steps.

To evaluate the influence of synaptic convolutional blocks and PMLIF models,
we design an ablation study of different strategies, which
consist of four: S0, SNNs with synaptic convolutional blocks and PMLIF; S1: SNNs with only PMLIF; S2: SNNs with only synaptic convolutional blocks; S3: SNNs without synaptic convolutional blocks and PMLIF. 
As shown in table \ref{T6}, The SNNs with Synaptic convolutional blocks and PMLIF have a higher testing accuracy and the performance improvement is more significant with the reduction of time steps.
Compared with the SNN without synaptic convolutional block and PMLIF, the SNN with synaptic convolutional block and PMLIF has a $1.38\%$ improvement on testing accuracy, when time step count is 2.

\begin{table}
\caption{Ablation study of different strategies based on DVS128-gesture }
\label{T6}
\centering
\resizebox{\columnwidth}{15mm}{
\begin{tabular}{lccccccccc} 
\hline
\multicolumn{1}{c}{\multirow{2}{*}{Methods}} & \multicolumn{3}{c}{T=2}        & \multicolumn{3}{c}{T=5}        & \multicolumn{3}{c}{T=10}        \\ 
\cline{2-10}
\multicolumn{1}{c}{}                         & Mean(\%) & Std(\%) & Best(\%) & Mean(\%) & Std(\%) & Best(\%) & Mean(\%) & Std(\%) & Best(\%)  \\ 
\hline
S0                                           & 95.60    & 0.40    & 96.52    & 98.47     & 0.31    & 98.96    & 98.75    & 0.28    & 98.96     \\
S1                                           &    95.35      &    0.17     &      95.49     &      98.06    &     0.52    &    98.96      & 98.33    & 0.38    & 98.61     \\
S2                                           &    94.79      &   0.31      &      95.14    &     97.99     &    0.40     &   98.61       & 98.13    & 0.28    & 98.61     \\
S3                                           &    94.65      &   0.36      &     95.14     &    97.98       &      0.24  &   98.26       & 97.92    & 0.22    & 98.26     \\
\hline
\end{tabular}
}
\end{table}

\subsection{Performance analysis}
\subsubsection{Influence of compression ratio}

\begin{figure}[t]
\subfigure[]{
\includegraphics[width=6cm]{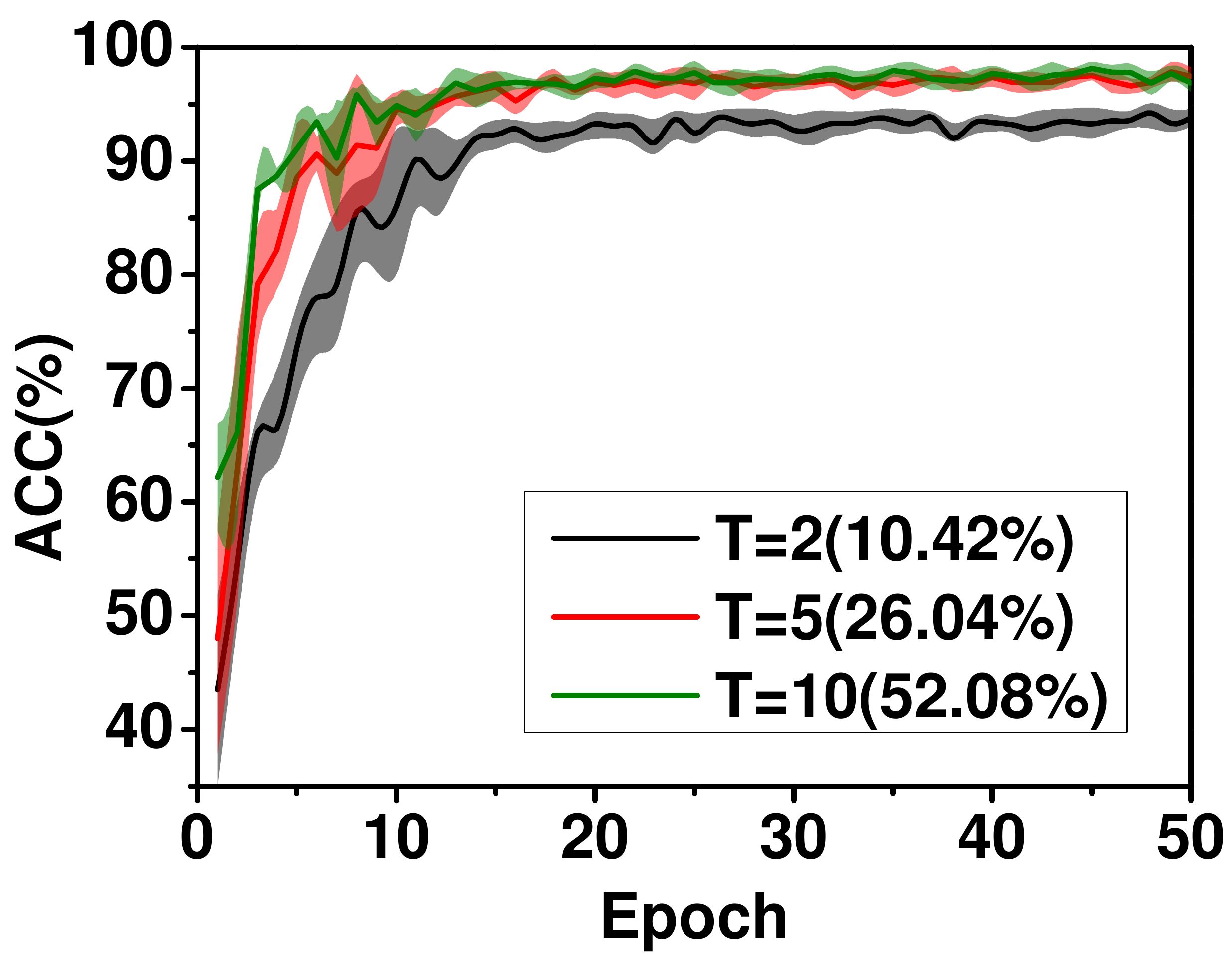}
%\label{DVS128}
}
\hspace{-6mm}
\subfigure[]{
\includegraphics[width=6cm]{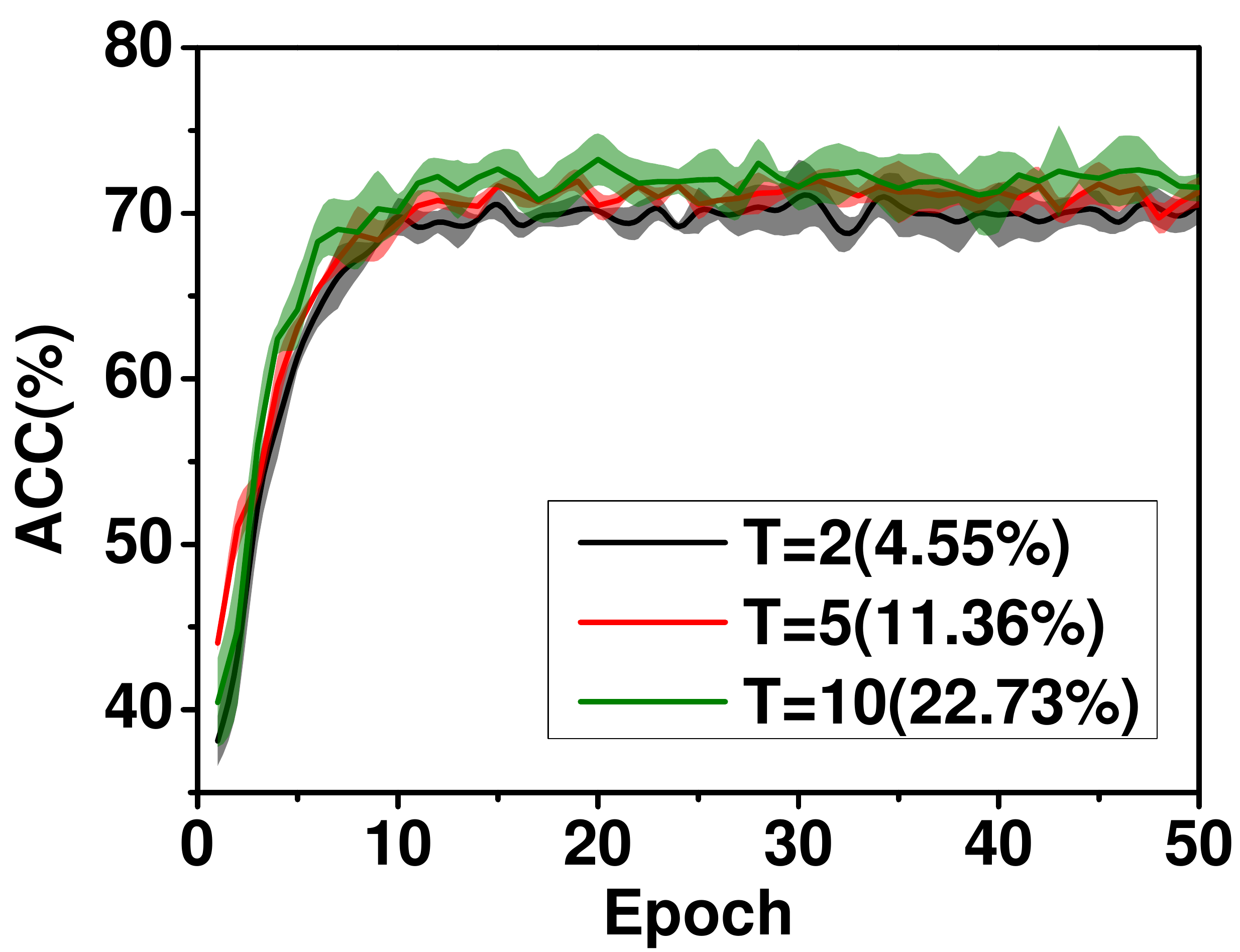}
%\label{Cifar10-DVS}
}
\caption{ The test accuracy of different compression ratios on (a) DVS128-gesture and  (b) Cifar10-DVS dataset}
\label{DVS128_Cifar10}
\end{figure}

To reduce the training and inference latency of SNNs,  we proposed a spatio-temporal compression method to aggregate event streams into a few time steps.
Since neuromorphic datasets will be integrated into tens or hundreds of frames in the reported works,
the compression ratio is defined as the ratio of average frames or time steps used in the mentioned work to the time steps we applied.
The lines in Fig. \ref{DVS128_Cifar10} denotes the mean of the accuracy obtained by repeating the experiments five times, the shade is the fluctuation range of the data.

 As Fig. \ref{DVS128_Cifar10} (a) is shown, the accuracy still can keep $98.96\%$ when compression ratio is $26.04\%$ for DVS128-gesture.
 When compression ratio is $10.42\%$, classification accuracy drops from $98.96\%$ to $95.83\%$.
 The reason for the significant accuracy degradation is the time steps are only two when the compression ratio is $10.42\%$, which causes too much temporal information to be lost.
 For Cifar10-DVS dataset, a higher compression ratio is applied.
 Compared with DVS128-gesture, the accuracy drop is not significant with the increase of the compression ratio. 
 The reason for that is Cifar10-DVS is a neuromorphic dataset converted from the static dataset, which contains less temporal information, high compression ratio mainly causes temporal information loss rather than spatial information.

\subsubsection{Distribution of $\tau_{mem}$ after training}
\begin{figure*}[t]
\centerline{\includegraphics[width=12cm]{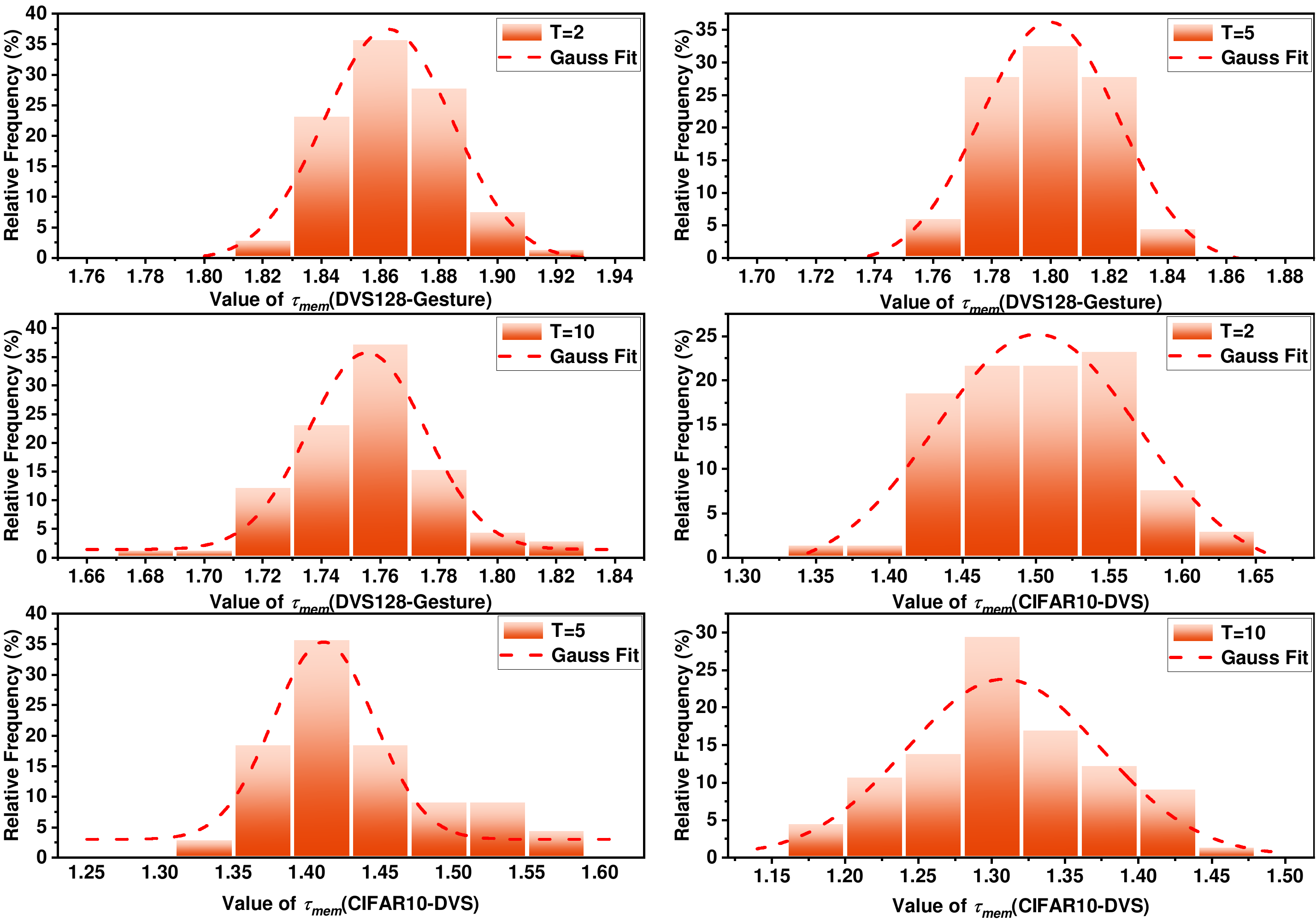}}
\caption{Distribution of $\tau_{mem}$  after training on DVS128-Gesture and CIFAR10-DVS}
\label{Distribution}
\end{figure*}

In PMLIF model, we introduce the learnable membrane constants to improve the convergence speed and stability of the whole network.
The results in section \ref{AblationStudy} have shown the benefit of applying the PMLIF model.
In this section, we study the distribution of the membrane constants under different compression ratios.
As Fig. \ref{Distribution} is shown, the distribution of membrane constants basically conformed to a Gaussian distribution.
We can find that the mean of membrane constants increases with the increase of the compression ratio.
The membrane time constants are used to control the retained information at the previous time step.
A larger membrane constant means more information at the previous time step will be retained.
The increase of compression ratio means that more information will be integrated into one time step.
The analysis result is that a larger membrane time constant should be applied to keep more information of previous time step retained when the compression ratio increases.

\section{Conclusion}
In this work, we proposed an ultra-low latency spiking neural network with spatio-temporal compression and a Synaptic convolutional block for event stream classification.
The proposed spatio-temporal compression method is used to compress event streams into few time steps to reduce the training and inference latency.
We also proposed a synaptic convolutional block as the first layer of the SNN, in which a synaptic layer is applied to balance the dramatic change between adjacent time steps.
The parametric multi-threshold Leaky Integrate-and-Fire models, whose membrane constants are learnable, are introduced in our SNNs.
We evaluate our proposed method and compare it with state-of-the-art methods.
Experiment results show that our proposed method outperforms state-of-the-art methods with the same or similar network architecture on neuromorphic datasets, such as N-MNIST, CIFAR10-DVS, DVS128-gesture.

\section*{Acknowledgment}

This work was supported in part by the National Natural Science Foundation of China under Grant 62004146, by the China Postdoctoral Science Foundation funded project under Grant 2021M692498, and by the Fundamental Research Funds for the Central Universities.

\bibliography{mybibfile}

\end{document}